\documentclass[letterpaper]{article} 
\usepackage{aaai2026} 
\usepackage{times}  
\usepackage{helvet}  
\usepackage{courier}  
\usepackage[hyphens]{url}  
\usepackage{graphicx} 
\usepackage{natbib}  
\usepackage{caption} 
\usepackage{booktabs} 
\usepackage{amsmath}
\usepackage{amssymb}
\usepackage{enumitem}
\usepackage{multirow}
\usepackage{xcolor}
\usepackage{algorithm}
\usepackage{algpseudocode}  

\usepackage{newfloat}
\usepackage{listings}
\DeclareCaptionStyle{ruled}{labelfont=normalfont,labelsep=colon,strut=off} 
\floatstyle{ruled}
\newfloat{listing}{tb}{lst}{}
\floatname{listing}{Listing}
\author{
   Bowei He\textsuperscript{\rm 1}, Bowen Gao\textsuperscript{\rm 2}, Yankai Chen\textsuperscript{\rm 3}, Yanyan Lan\textsuperscript{\rm 2, 4}\footnotemark[1], Chen Ma\textsuperscript{\rm 1}\thanks{Corresponding Authors: lanyanyan@air.tsinghua.edu.cn and chenma@cityu.edu.hk.}, \\ Philip S. Yu\textsuperscript{\rm 3}, Ya-Qin Zhang\textsuperscript{\rm 2}, Wei-Ying Ma\textsuperscript{\rm 2}
}
\affiliations{
    \textsuperscript{\rm 1} City University of Hong Kong,  \textsuperscript{\rm 2} Institute for AI Industry Research (AIR), Tsinghua University, \\ \textsuperscript{\rm 3} University of Illinois Chicago, \textsuperscript{\rm 4} Beijing Academy of Artificial Intelligence
}

\title{S$^2$Drug: Bridging Protein Sequence and 3D Structure in Contrastive Representation Learning for Virtual Screening}

\usepackage{bibentry}

\begin{document}

\maketitle

\begin{abstract}
Virtual screening (VS) is an essential task in drug discovery, focusing on the identification of small-molecule ligands that bind to specific protein pockets. Existing deep learning methods, from early regression models to recent contrastive learning approaches, primarily rely on structural data while overlooking protein sequences, which are more accessible and can enhance generalizability. 
However, directly integrating protein sequences poses challenges due to the redundancy and noise in large-scale protein-ligand datasets.
To address these limitations, we propose \textbf{S$^2$Drug}, a two-stage framework that explicitly incorporates protein \textbf{S}equence information and 3D \textbf{S}tructure context in protein-ligand contrastive representation learning. In the first stage, we perform protein sequence pretraining on ChemBL using an ESM2-based backbone, combined with a tailored data sampling strategy to reduce redundancy and noise on both protein and ligand sides. In the second stage, we fine-tune on PDBBind by fusing sequence and structure information through a residue-level gating module, while introducing an auxiliary binding site prediction task. This auxiliary task guides the model to accurately localize binding residues within the protein sequence and capture their 3D spatial arrangement, thereby refining protein-ligand matching.
Across multiple benchmarks, S$^2$Drug consistently improves virtual screening performance and achieves strong results on binding site prediction, demonstrating the value of bridging sequence and structure in contrastive learning.
\end{abstract}

\section{Introduction}
Virtual screening (VS) is regarded as a cornerstone of modern drug discovery, as it enables researchers to computationally evaluate large libraries of chemical compounds to identify potential drug candidates that bind to a target protein~\citep{shoichet2004virtual}. By leveraging high-performance computation algorithms and hardwares, VS significantly reduces the cost, time, and labor required in early-stage hit identification, especially when compared to traditional high-throughput empirical screening over trillions of or even more drug-like molecules~\citep{lyu2023modeling}.

Molecular docking and deep learning methods are two types of the most mainstream approaches for virtual screening. The docking-based methods~\citep{friesner2004glide} simulates the physical binding process between a ligand and a protein to estimate binding affinity, while learning-based methods leverage data-driven models to predict interactions without explicit simulation. 
Notably, the success of deep learning in related fields such as graph learning and natural language processing has fueled growing interest in learning-based approaches. Early works in this category consider the VS as a binding affinity regression or classification task~\citep{wu2022bridgedpi, zhang2023planet}. Recent works like DrugCLIP~\citep{gao2023drugclip} and DrugHash~\citep{han2025drughash} model the VS as a retrieval task, where protein pockets are regarded as queries to retrieve matching ligands. To enable this, they leverage contrastive learning to align protein and ligand representations by pulling binding pairs closer and pushing apart non-binding pairs in the embedding space. 
However, whether molecular docking or existing learning methods, most approaches still rely exclusively on 3D structures~\citep{maia2020structure}, neglecting the employment of protein sequence information. 
In fact, this reliance on single-conformer atom-level structural information can make models overly sensitive to input perturbations, and less robust to pocket conformational flexibility-a common challenge in practical drug discovery. 
Furthermore, determining protein 3D structures requires techniques such as X-ray crystallography and Cryo-EM, which are time-consuming and costly. These limitations hinder the expansion of large-scale training datasets, ultimately restricting the performance and generalizability of VS methods.

The protein research in recent decades follows a fundamental principle: \textit{"Sequence determines structure, and structure determines function"}~\citep{anfinsen1973principles}. This implies that amino acid sequences fundamentally encode the information necessary for protein folding and function, which in turn influences how proteins interact with ligands, especially their affinity to ligands in specific sites. 
Early works~\citep{shen2023svsbi} have explored the introduction of protein sequence information for VS. 
However, they merely leverage sequence representations encoded by pretrained protein LSTM~\citep{alley2019unified} or Transformer~\citep{rives2021biological} models, without explicitly learning the protein sequence-ligand interactions. 
Although large-scale protein sequence–ligand interaction datasets such as ChemBL~\citep{gaulton2012chembl} have been available for years, their potential to directly supervise the learning of sequence-level binding preferences remains underexplored.
Nevertheless, directly integrating such datasets into VS model training poses significant challenges. Because the data redundancy and noise in both protein and ligand sides-such as homology/functional redundancy, binding affinity variability, and non-specific binders-may significantly hinder the model performance and generalizability. Moreover, relying solely on protein sequence information while discarding structural context severely limits the model's ability to identify binding pockets, model spatial interactions (e.g., shape complementarity), and account for conformational flexibility. 
These factors are critical for accurate and generalizable virtual screening.

To tackle such challenges, we propose \textbf{S$^2$Drug}, a two-stage contrastive learning framework that bridges protein \textbf{S}equence and 3D \textbf{S}tructure information to enable more accurate virtual screening. 
(1) In the first stage, we perform protein sequence-ligand contrastive representation pretraining on the large-scale dataset ChemBL. During this stage, we design a bilateral data-sampling strategy: on the protein side, we apply homology-aware downweighting and functional deduplication to reduce protein-side redundancy; on the ligand side, we apply affinity variability filtering and frequent hitter removal to mitigate noise. (2) In the next stage, we introduce a residue-level gating module to fuse protein sequence and 3D structure information. This enables fine-tuning of the learned representations on the small-scale PDBBind dataset, which provides high-resolution pocket structures.
In addition, we introduce an auxiliary binding site prediction task that determines whether each residue belongs to the binding site. Given that a pocket is essentially the spatial cavity-shape aggregation of binding site residues scattered across the sequence, this task enhances the model's understanding of protein 3D folding-particularly in pocket regions-thereby further improving VS performance.

We summarize our contributions as follows:
\begin{itemize}[leftmargin=*, topsep=2pt]
    \item We propose a two-stage contrastive learning framework, namely S$^2$Drug, which performs sequence pretraining and sequence-structure fusion finetuning for virtual screening.
    \item We devise a data sampling strategy to harness large-scale protein sequence-ligand datasets, ensuring high-quality pretraining data by reducing redundancy and noise.
    \item We incorporate a binding site prediction auxiliary task in the fine-tuning stage to enhance model's understanding of binding pocket locations and their spatial configurations.
    \item We evaluate our S$^2$Drug framework on two most typical virtual screening datasets and obtain consistent superior performance over baselines. Additionally, S$^2$Drug exhibits compelling results on binding site prediction.
\end{itemize}

\section{Related Works}
\paragraph{Virtual Screening} Existing methods for virtual screening can be broadly categorized into two main approaches: docking-based and learning-based methods. 
Docking-based methods, such as Glide-SP \citep{friesner2004glide} and Vina \citep{trott2010autodock}, is a computational technique that predicts how a small molecule (ligand) binds to a target protein’s binding site, estimating both the binding pose and interaction strength. However, this type of methods are often time-consuming because they must search numerous ligand conformations and evaluate each with computationally expensive scoring functions. On the other hand, learning-based methods leverage machine learning and deep learning techniques to predict ligand-protein pocket interactions more efficiently. Within learning-based methods, there are three primary subcategories: regression~\citep{ozturk2018deepdta, zheng2019onionnet, zhang2023planet}, classification~\citep{wu2022bridgedpi, yazdani2022attentionsitedti}, and retrieval~\citep{gao2023drugclip, han2025drughash} approaches. Among them, while structural data has been extensively used, the potential of protein sequence data, which is more readily available and can provide complementary information, has been largely neglected.

\paragraph{Protein Representation Learning} Though the importance of molecular representation learning in drug discovery tasks has been widely recognized~\citep{wang2022molecular}, the most advanced protein representation learning methods in the virtual screening still stay in the structure learning stage~\citep{han2025drughash}. They utilize pretrained SE(3) Transformers to model 3D atom-level interactions, and adopt contrastive learning to align protein and ligand representations.
In contrast, out of virtual screening domain, the protein sequence representation learning has achieved great progress in recent years due to the widely available sequence data, low annotation cost without structure determination, and the model architecture derived from language models~\citep{lin2023evolutionary, abramson2024accurate}. Besides, there is also increasing interest in incorporating both sequence and structure information to conduct more comprehensive protein representation learning~\citep{lee2023pre, hu2024learning, li2024improving}. 
However, how to employ protein sequence representation learning in virtual screening, especially sequence-structure collaborative learning, is still less explored, which is the focus of this work.

\section{Methodology}
\begin{figure*}[t]
    \centering
    \includegraphics[width=0.85\textwidth]{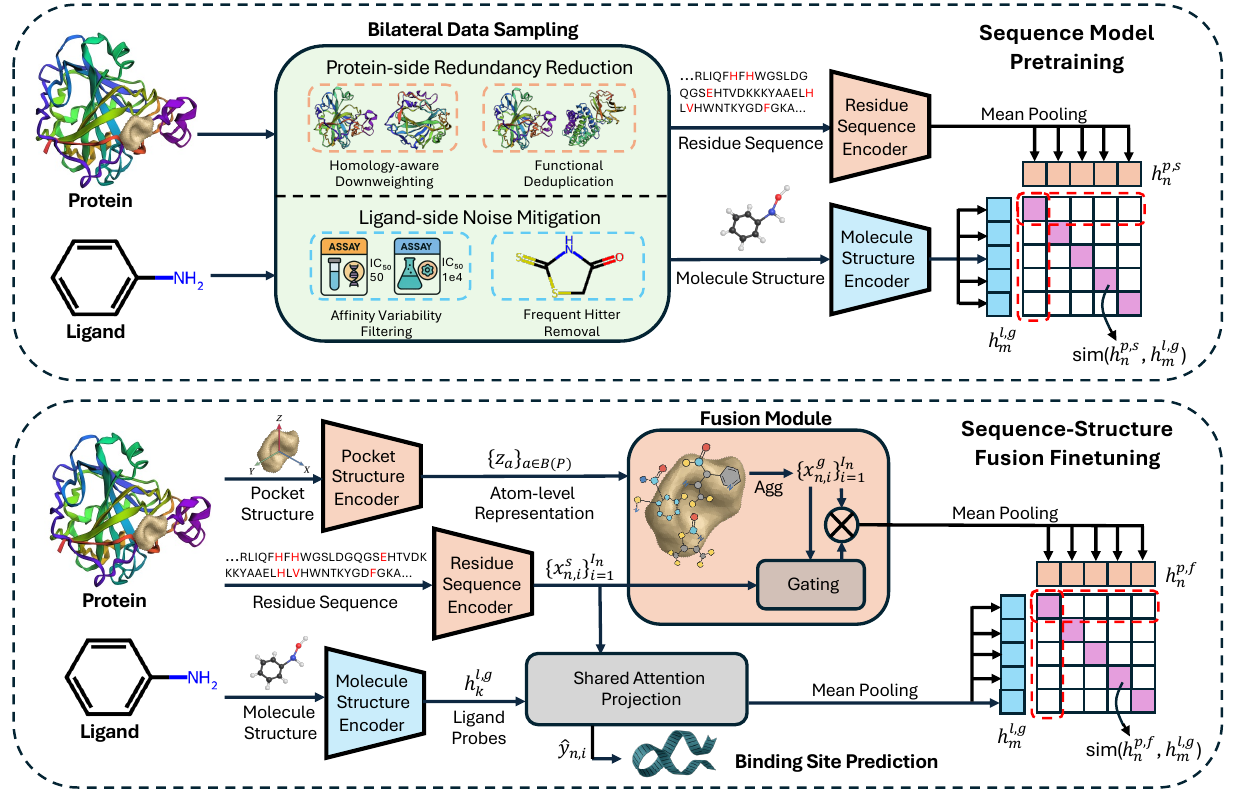}
    \vspace{-2mm}
    \caption{The illustration of our proposed two-stage contrastive representation learning framework S$^2$Drug for bridging protein sequence and 3D structure. The red characters indicate pocket residues, which are however agnostic to the sequence encoder.}
   \label{fig: overall}
    \vspace{-3mm}
\end{figure*}

In this section, we first define the problem, and then detail the two stages of our S$^2$Drug framework: (1) sequence model pretraining and (2) sequence-structure fusion finetuning.
We illustrate the framework in Figure~\ref{fig: overall}.

\subsection{Problem Formulation}
In this work, we define two interrelated tasks to enhance virtual screening in drug discovery. Note we use the term ``ligand” to refer to candidate small molecules, regardless of whether binding has been experimentally confirmed.  

\paragraph{Main Task: Virtual Screening} 
The primary objective is to predict the binding likelihood between a specific protein binding pocket and a ligand, enabling the identification of potential drug candidates from a ligand library. Given a protein $P$, which includes its amino acid sequence $S(P)$ and three-dimensional structural information $G(P)$, we denote its binding pocket as $B(P) \subseteq P$, defined by a subset of residues and their spatial configuration. For a ligand $L$, represented by its molecular structure $G(L)$, we aim to learn a scoring function $s(B(P), L): \mathcal{B} \times \mathcal{L} \to \mathbb{R}$, where $\mathcal{B}$ is the space of protein binding pockets, and $\mathcal{L}$ is the space of ligands. The function $s(B(P), L)$ outputs a real-valued score indicating the likelihood of ligand $L$ binding to the pocket $B(P)$, which is used to rank ligands for a given protein target based on their affinity to the specified binding pocket.

\paragraph{Auxiliary Task: Binding Site Prediction} 
The auxiliary task focuses on identifying the amino acid residues in the protein sequence that constitute the binding site, thereby enhancing the performance of the main task. For a given protein $P$, its sequence is represented as $S(P) = (r_1, r_2, \dots, r_{I})$, where $r_i$ is the $i$-th amino acid residue, and $I$ is the sequence length. We predict a binary label $y_i \in \{0,1\}$ for each residue $s_i$, where $y_i = 1$ indicates that the residue is part of the binding site, and $y_i = 0$ indicates otherwise. This task is formulated as a sequence labeling problem, where the model outputs a prediction vector $\hat{Y}(P) = (\hat{y}_1, \hat{y}_2, \dots, \hat{y}_I)$, with $\hat{y}_i$ representing the predicted probability that residue $i$ belongs to the binding site.

By jointly optimizing these two tasks, the model learns representations that integrate both sequence and structural information, thereby improving the accuracy and effectiveness of virtual screening.

\subsection{Sequence Model Pretraining}
\label{sec:pretraining}

Although large-scale protein language models have demonstrated strong  performance across a wide range of tasks, they are primarily trained on general sequence corpora without explicit supervision from protein–ligand interactions. To effectively model residue-level representations that are sensitive to binding specificity, it is necessary to further adapt these sequence models using protein–ligand interaction data.

\subsubsection{Bilateral Data Sampling}
To ensure high-quality and generalizable supervision from large-scale protein–ligand interaction datasets such as ChemBL, we introduce a holistic strategy that jointly filters and subsamples the training data from both the protein and ligand sides. Let $\mathcal{D}_0 = \{(P_n, L_n, a_n)\}^{|\mathcal{D}_0|}_1$ denote the initial dataset with 745K protein–ligand–affinity triplets, where $P_n$ is the protein with $S(P_n)$ as corresponding residue sequence, $L_n$ is the ligand, and $a_n \in \mathbb{R}$ is the reported binding affinity. We construct a cleaned subset $\mathcal{D} \subset \mathcal{D}_0$ via the following two-sided strategy: 

\textbf{Step 1: Protein-side Redundancy Reduction} 
Protein data in ChemBL suffers from severe homology and functional redundancy, which may encourage the memorization of family-specific patterns rather than the learning generalizable interaction rules.

\textit{(1) Homology-aware Downweighting.} We compute pairwise residue sequence similarity $\mathrm{SeqSim}(S(P_n), S(P_m)) \in [0,1]$, based on normalized sequence alignment scores. Then we cluster the sequences using MMseqs2 at a 40\% identity threshold~\citep{kallenborn2024gpu}, yielding clusters $\mathcal{C}^{hom} = \{C_1^{hom}, \dots, C_M^{hom}\}$. The sampling probability of each protein $P_n \in C_m^{hom}$ is defined as:
\begin{equation}
\Pr(P_n) = \frac{1}{|C_m^{hom}|^\alpha}, \quad \alpha \in (0,1].
\end{equation}
We set $\alpha = 0.5$ to penalize large protein families while retaining sufficient diversity.

\textit{(2) Functional Deduplication.} Functional isoforms or mutants may be overrepresented, contributing negligible additional information. Based on UniProt or Gene Ontology function annotations $\phi(P_i)$, we define groups:
\begin{equation}
C^{fun}_k = \{P_n \mid \phi(P_n) = \phi_k \}.
\end{equation}
We retain a single representative per $C^{fun}_k$, prioritizing proteins with greater ligand diversity.

\textbf{Step 2: Ligand-side Noise Mitigation}
The noise in the ligand side arises from binding affinity variability across assays and from non-specific promiscuous compounds.

\textit{(1) Affinity Variability Filtering.} For duplicate protein-ligand pairs with different affinity values obtained under different assay conditions $\{(P_n, L_n, a_n^j\}_{j=1}^{J}$, we calculate:
\begin{equation}
\sigma_n = \mathrm{StdDev}(\log a_n^1, \dots, \log a_n^J).
\end{equation}
Only instances with $\sigma_n < \delta$ (we use $\delta = 1.0$) are retained, and the canonical affinity is set as the mean log-value.

\textit{(2) Frequent Hitter Removal.} Some compounds exhibit promiscuous binding activity across multiple targets, not due to specific molecular recognition but rather because of reactive substructures. These artifacts can mislead representation learning.  Let $f(L_n)$ be the number of proteins a ligand binds. We discard ligands where $f(L_n) > T$ (with $T = 20$) unless a majority of affinities indicate strong binding, i.e., high affinity values. Besides, ligand molecules with known such reactive substructures (e.g., PAINS) are filtered out.

\textbf{Step 3: Joint Subsampling with Rebalancing} After cleaning, we subsample remaining data with rebalancing:
\begin{equation}
\mathcal{D} = \mathrm{Sample}_{(P, L, a) \sim \mathcal{D}_0} \left[ \Pr(P) \cdot \mathbb{I}_{\mathrm{clean}(P,L,a)} \cdot w_{\mathrm{lig}}(L) \right],
\end{equation}
where $\mathbb{I}_{\mathrm{clean}(P,L,a)} \in \{0,1\}$ indicates that a triplet passes all filtering rules, and $w_{\mathrm{lig}}(L) \propto 1/f(L)$ downweights high-frequency ligands.

\subsubsection{Representation Learning} 
To learn modality-specific representations that simultaneously capture protein sequence semantics and ligand structural properties, we first employ modality-aligned encoders for both proteins and ligands. Then, a symmetric contrastive training objective is applied to align their embeddings in a shared latent space.

\textbf{Sequence Encoder for Proteins} We adopt the pretrained ESM2 model~\cite{lin2023evolutionary} to encode the amino acid residue sequences. Given a protein $P_n$ with its residue sequence $S(P_n)=(r_1, \dots, r_I)$, the sequence encoder $\text{Seq}^p(\cdot)$ maps it to a fixed-length embedding $h_n^{p,s} \in \mathbb{R}^{d_s}$:
\begin{equation}
h_n^{p,s} = \text{MeanPool}(\text{Seq}^p(S(P_n))),
\end{equation}
where $\text{Seq}^p(\cdot)$ is initialized from the 650M-parameter ESM2 model. The output $h_n^{p,s}$ is the mean pooling of tokens' representation from the final Transformer layer.

\textbf{Structure Encoder for Ligands} For ligand molecules, we adopt the Uni-Mol molecular encoder~\cite{zhouuni}, which encodes 3D conformers using atom types and pairwise Euclidean distances. Given a ligand $L_i$ represented by structual information $G(L_n)$ including 3D atomic coordinates and associated atom types, the encoder $\text{Stru}^l(\cdot)$ outputs an embedding $h_n^{l,g} \in \mathbb{R}^{d_g}$ after mean pooling over atoms:
\begin{equation}
h_n^{l,g} = \text{MeanPool}(\text{Stru}^l(G(L_n))).
\end{equation}

To align its representation space with the protein sequence encoder, we project both $h_n^{p,s}$ and $h_n^{l,g}$ into a shared embedding space via two-layer MLPs with layer normalization.

\textbf{Contrastive Training Objective} We train the protein and ligand encoders using a symmetric contrastive objective that brings matching pairs closer while pushing apart non-matching ones. For a batch of $N$ protein–ligand pairs $\{(P_n, L_n)\}_{n=1}^{N}$, we compute the pairwise cosine similarity: 
\begin{equation}
\mathrm{sim}(h_n^{p,s}, h_m^{l,g}) = \frac{\langle h_n^{p,s}, h_m^{l,g} \rangle}{\|h_n^{p,s}\| \cdot \|h_m^{l,g}\|}.
\end{equation}
The contrastive loss is defined via InfoNCE~\citep{oord2018representation}:
\begin{equation}
\begin{aligned}
\mathcal{L}_{\text{pc}} = -\frac{1}{N} \sum_{n=1}^N & \log \frac{\exp(\mathrm{sim}(h_n^{p,s}, h_n^{l,g})/\tau)}{\sum_{m=1}^N \exp(\mathrm{sim}(h_n^{p,s}, h_m^{l,g})/\tau)} + \\ 
&\log \frac{\exp(\mathrm{sim}(h_n^{p,s}, h_n^{l,g})/\tau)}{\sum_{m=1}^N \exp(\mathrm{sim}(h_m^{p,s}, h_n^{l,g})/\tau)},
\label{equ:pretrain contrastive}
\end{aligned}
\end{equation}
where $\tau$ is a temperature hyperparameter controlling the sharpness of the similarity distribution. 

\subsection{Sequence-Structure Fusion Finetuning}
To enhance virtual screening accuracy, it is crucial for the model to incorporate both the evolutionary information embedded at protein sequences and the fine-grained geometric context provided by 3D pocket structures. While structural encoders are effective in modeling atomic interactions, they often suffer from sensitivity to conformational perturbations and may not generalize well to novel proteins. To address this, we propose a fusion-based finetuning strategy that combines above pretrained protein sequence embeddings with structure-aware representations. They are jointly trained using a contrastive interaction objective and an auxiliary binding site prediction task on the PDBBind dataset.

\subsubsection{Sequence-Structure Fusion Module}
Given a protein $P_n$, we obtain residue-level sequence representation $\{x_{n,i}^s\}_{i=1}^{I_n}$ from the encoder $\text{Seq}^p$ and the atom-level structural representation $\{z_a\}_{a \in B(P)}$ from pocket structure encoder $\text{Stru}^{p}$ initialized from Uni-Mol. For each pocket residue $r_i \subset B(P)$, we perform masked mean pooling over its constituent atoms to aggregate the structure-based embedding:
\begin{equation}
x^g_{n,i} = \frac{1}{|r_i|} \sum_{a \in r_i} \mathbb{I}_{a \in r_i} \cdot z_a,    
\end{equation}
where $\mathbb{I}_{a \in r_i}$ indicates the atom $a$ is in the pocket residue $r_i$. We then apply a \textit{gating} mechanism to adaptively fuse the sequence and structure information, thus obtaining $x^f_{n,i}$:
\begin{equation}
\begin{aligned}
\beta_{n,i} &= \sigma\left(W_\beta^\top [W_s x^s_{n,i}; W_g x^g_{n,i}] + b_\beta\right), \\
x^f_{n,i} &= \beta_{n,i} \cdot W_s x^s_{n,i} + (1 - \beta_{n,i}) \cdot W_g x^g_{n,i}, 
\end{aligned}
\end{equation}
where $\sigma(\cdot)$ denotes the sigmoid function, and $W_\beta$, $b_\beta$ are learnable parameters. Also, $W_s, W_g$ are learnable matrices to project $x^s_{n,i}$ and $x^g_{n,i}$ into shared representation space. This fusion mechanism allows the model to dynamically prioritize the most informative modality for each residue, depending on the local sequence-structure alignment. Finally, after two Transformer layers, we conduct mean pooling over pocket residues to obtain fused pocket representation $h_n^{p,f}$.

\subsubsection{Auxiliary Binding Site Prediction}
As we know, a pocket is a spatial aggregation of residues in the three-dimensional conformation, rather than a continuous segment in the primary sequence. These residues come together to form a binding cavity as a result of protein folding that brings them into close proximity in 3D space. Therefore, conducting binding site prediction on the entire protein sequence helps the model infer its spatial arrangement-particularly the conformation and further biochemical properties of the pocket region-ultimately facilitating the accurate virtual screening process.
To obtain the residue distribution of the binding pocket on the whole residue sequence, for each protein, we sample $K$ ligand probes $\{L_k\}_{k=1}^{K}$ and compute their interaction relevance with residues via a \textit{shared attention projection}:
\begin{equation}
\alpha_{n,i}^k = \frac{\exp(W_rx^s_{n,i} \cdot W_l h^{l,g}_k)}{\sum_{i=1}^{I_n} \exp(W_rx^s_{n,i} \cdot W_l h^{l,g}_k)},
\end{equation}
where $h^{l,g}_k$ is the structural representation of the $k$-th probe ligand from $\text{Stru}^l$, and $W_r, W_l$ are trainable projection matrices. 
To avoid information leakage, we only utilize the sequence representation $x^s_{n,i}$ here. We then average the attention scores to obtain the per-residue binding probability:
\begin{equation}
\hat{y}_{n,i} = \frac{1}{K} \sum_{k=1}^K \alpha^k_{n,i}.
\end{equation}
The prediction is trained with summed binary cross-entropy:
\begin{equation}
\mathcal{L}_{\text{bsp}} = -\frac{1}{N} \sum_{n=1}^{N} \sum_{i=1}^{I_n} \left[ y_{n,i} \log \hat{y}_{n,i} + (1 - y_{n,i}) \log (1 - \hat{y}_{n,i}) \right].
\end{equation}

\subsubsection{Training Objective} To align protein–ligand interactions while emphasizing spatially grounded residue position, we jointly optimize a dual loss objective. For contrastive training, we encode a protein pocket into $h^{p,f}_n$ via pooling over fused residue embeddings, and a ligand into $h^{l,g}_n$ using the $\text{Stru}^l$ encoder. The contrastive loss is defined as:
\begin{equation}
\begin{aligned}
\mathcal{L}_{\text{fc}} = -\frac{1}{N} \sum_{n=1}^N
&\log \frac{\exp(\text{sim}(h^{p,f}_n, h^{l,g}_n) / \tau)}{\sum_{m=1}^N \exp(\text{sim}(h^{p,f}_n, h^{l,g}_m) / \tau)} + \\
&\log \frac{\exp(\text{sim}(h^{p,f}_n, h^{l,g}_n) / \tau)}{\sum_{m=1}^N \exp(\text{sim}(h^{p,f}_m, h^{l,g}_n) / \tau)}, 
\label{equ:finetune contrastive}
\end{aligned}
\end{equation}
where $\text{sim}(\cdot, \cdot)$ denotes cosine similarity, $N$ is the batch size, and $\tau$ is the temperature parameter, similar to pretraining phase.
The final training objective combines two losses:
\begin{equation}
\mathcal{L}_{\text{total}} = \mathcal{L}_{\text{fc}} + \lambda \cdot \mathcal{L}_{\text{bsp}},
\label{equ:combined}
\end{equation}
where $\lambda$ is a balancing coefficient. This learning scheme enables the model to simultaneously learn global interaction alignment and local spatial specificity, improving both virtual screening accuracy and binding site interpretability.


\section{Experiment}
\subsection{Evaluation Settings}
\paragraph{Datasets} We evaluate the virtual screening performance of various methods on the DUD-E~\citep{mysinger2012directory} and LIT-PCBA~\citep{tran2020lit}. As for the binding site prediction task,  HOLO4K, COACH420~\citep{krivak2018p2rank}, ASD~\citep{liu2020unraveling} are utilized as the benchmarks. To ensure fairness, all the evaluations in this paper follow the \textit{zero-shot} setting.

\paragraph{Baselines} For virtual screening evaluation, we take several shared baselines over above two datasets: \textit{docking}: Glide-SP~\citep{friesner2004glide}, \textit{learning}: Planet~\citep{zhang2023planet}, Banana~\citep{brocidiacono2023bigbind}, SVSBI~\citep{shen2023svsbi}, DrugCLIP~\citep{gao2023drugclip}, DrugHash~\citep{han2025drughash}. Besides, we also take specialized baselines on DUD-E: \textit{docking}: Vina~\citep{trott2010autodock}, \textit{learning}: NN-score~\citep{durrant2011nnscore}, RF-score~\citep{ballester2010machine}, Pafnucy~\citep{stepniewska2017pafnucy}. On LIT-PCBA, we compare with following exclusive baselines: \textit{docking}: Surflex~\citep{spitzer2012surflex}, \textit{learning}: DeepDTA~\citep{ozturk2018deepdta}, Gnina~\citep{mcnutt2021gnina}. As for the evaluation of binding site prediction, we adopt P2Rank~\citep{krivak2018p2rank}, VN-EGNN~\citep{sestak2023vn}, DiffDock~\citep{corso2023diffdock}, and VN-EGNNrank~\citep{schneckenreiter2025ligand} as the baselines. Given these methods directly predict the pocket center coordinates, residues whose heavy atoms are within a fixed distance threshold (8 \text{\AA}) from the predicted center are considered as binding site residues.

\paragraph{Metrics} Following previous works~\citep{gao2023drugclip, han2025drughash}, we utilize the following metrics to evaluate virtual screening accuracy: the area under the receiver operating characteristic curve (AUROC), the Boltzmann-enhanced discrimination of receiver operating characteristic (BEDROC), enrichment factor (EF). For binding site prediction, we take F1 Score and the area under the precision-recall curve (PR-AUC), which is specially suitable for this task due to the extremely unbalanced labels. All reported metrics are scaled to percentage values for presentation clarity.

\paragraph{Implementation Details}
We conduct our experiments on a local server with 1.6 TB RAM, 8 $\times$ NVIDIA RTX A6000 GPUs, and dual Intel Xeon Silver 4309Y processors (16 cores, 32 threads @ 2.80 GHz). The epoch, learning rate, and batch size for pretraining stage are set as 10, 1e-3, and 128, respectively. During the finetuning stage, the epoch and batch size values are set to 50 and 48, while the learning rate keeps unchanged. As for the parameter initialization of sequence encoder and structure encoders, we employ the 650M version of ESM2 pretrained model and the 181MB version of Uni-Mol pretrained models, respectively. Statistical significance is assessed using paired t-tests, with performance improvements over best baselines considered significant at $p < 0.01$ (marked by $^*$ in tables). We run each experiment for 5 times and report the average results. The code has been provided in supplementary materials.


\subsection{Results and Analysis}
\begin{table}[h]
\centering
\resizebox{0.48\textwidth}{!}{
\begin{tabular}{l|c|c|c|c|c}
\toprule
& AUROC & BEDROC & EF$^{0.5\%}$ & EF$^{1\%}$ & EF$^{5\%}$ \\
\midrule
Glide-SP & 76.70 & 40.70 & 19.39 & 16.18 & 7.23 \\
Vina & 71.60 & - & 9.13 & 7.32 & 4.44 \\
\midrule
NN-score & 68.30 & 12.20 & 4.16 & 4.02 & 3.12 \\
RF-Score & 65.21 & 12.41 & 4.90 & 4.52 & 2.98 \\
Pafnucy & 63.11 & 16.50 & 4.24 & 3.86 & 3.76 \\
Planet & 71.60 & - & 10.23 & 8.83 & 5.40 \\
Banana & 50.14 & 2.40 & 1.19 & 1.18 & 1.01 \\
SVSBI & 76.28 & 42.35 & 25.64  & 12.82 & 7.03 \\
DrugCLIP & 79.45 & 47.82 & 37.86 & 30.76 & 10.10 \\
DrugHash & 83.73 & 57.16 & 43.03 & 37.18 & 12.07 \\
\midrule
S$^2$Drug & \textbf{92.46*} & \textbf{79.25*} & \textbf{58.37*} & \textbf{43.06*} & \textbf{18.82*} \\
\bottomrule
\end{tabular}}
\caption{Virtual screening performance comparison of different methods on DUD-E. The bold indicates best result.}
\label{tab:vs dude}
\end{table}

\begin{table}[h]
\centering
\resizebox{0.48\textwidth}{!}{
\begin{tabular}{l|c|c|c|c|c}
\toprule
& AUROC & BEDROC & EF$^{0.5\%}$ & EF$^{1\%}$ & EF$^{5\%}$ \\
\midrule
Surflex & 51.47 & - & - & 2.50 & - \\
Glide-SP & 53.15 & 4.00 & 3.17 & 3.41 & 2.01 \\
\midrule
DeepDTA & 56.27 & 2.53 & - & 1.47 & - \\
Planet & 57.31 & - & 4.64 & 3.87 & 2.43 \\
Gnina & 60.93 & 5.40 & - & 4.63 & - \\
Banana & \textbf{62.78} & 5.02 & 3.98 & 3.79 & 2.83 \\
SVSBI & 53.77 & 3.84 & 3.05 & 2.71 & 1.92 \\
DrugCLIP & 56.36 & 6.78 & 7.77 & 5.66 & 2.32 \\
DrugHash & 54.58 & 7.14 & 9.65 & 6.14 & 2.42 \\
\midrule
S$^2$Drug & 58.23 & \textbf{8.69*} & \textbf{11.44*}  & \textbf{7.38*} & \textbf{2.97*}\\
\bottomrule
\end{tabular}}
\caption{Virtual screening performance comparison of different methods on LIT-PCBA. The bold indicates best result.}
\vspace{-3mm}
\label{tab:vs pcba}
\end{table}

\paragraph{Overall Screening Performance}
To evaluate whether our proposed S$^2$Drug can effectively identify small-molecule ligands binding to the given protein pockets, we conduct experiments on DUD-E and LIT-PCBA datasets, with results shown in Tables~\ref{tab:vs dude} and \ref{tab:vs pcba}. From these results, we can observe that S$^2$Drug significantly outperforms all docking-based and learning-based baselines on all metrics. In particular, S$^2$Drug outperforms recent retrieval-based methods including DrugCLIP and DrugHash by a large margin (e.g., 13.01 and 8.73 points on AUROC, respectively), even though these methods also employ the contrastive representation learning. This demonstrates that our two-stage representation learning framework-designed to bridge protein sequence and 3D structure-can indeed improve VS accuracy.

\begin{figure}[h]
    \centering
    \includegraphics[width=0.5\textwidth]{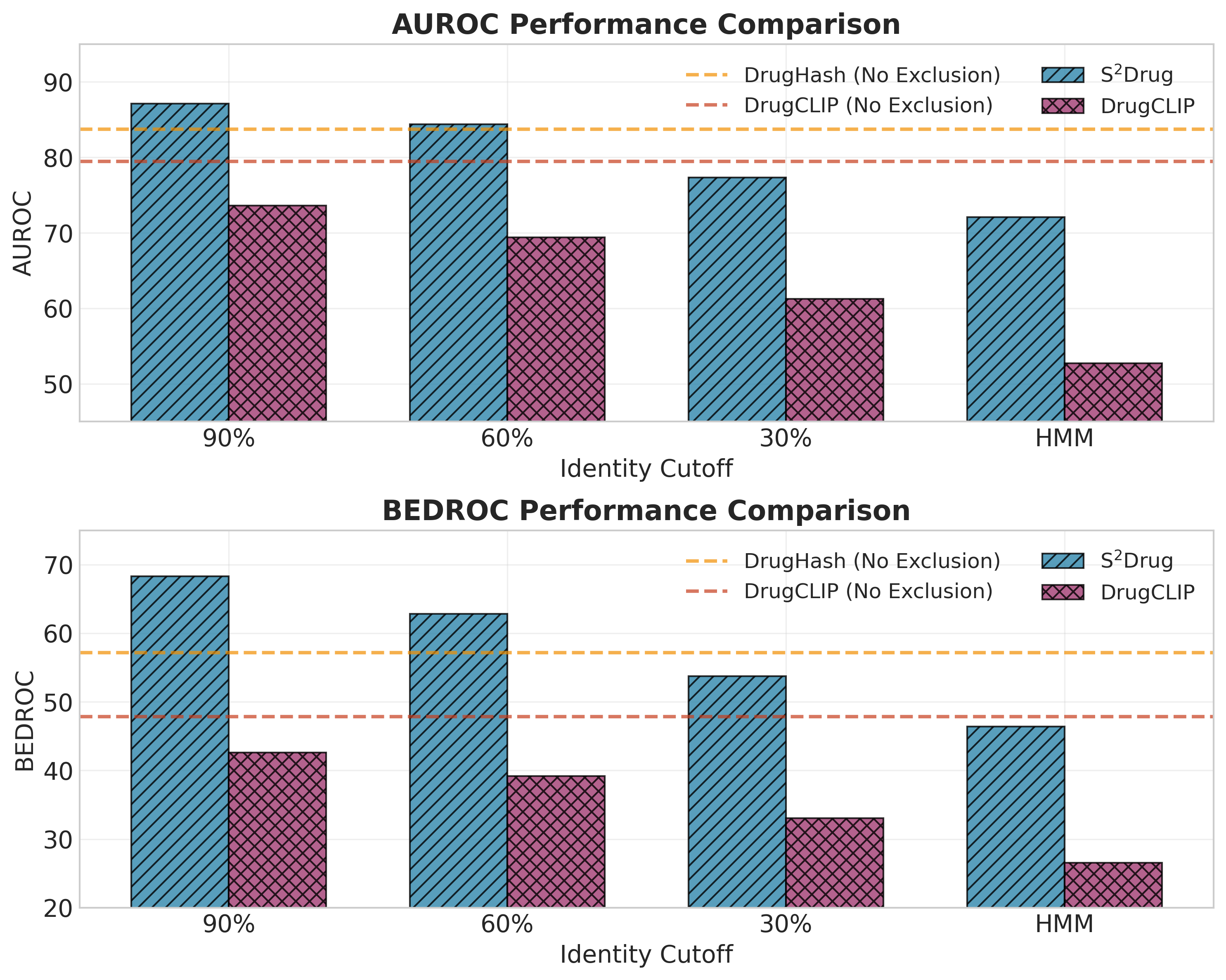}
    \caption{Virtual screening experiments on homology exclusion scenarios to evaluate method generalizability.}
   \label{fig: exclusion}
    \vspace{-3mm}
\end{figure}

\paragraph{Evaluation on Homology Exclusion Scenarios}
To assess the S$^2$Drug’s generalization ability, we conduct the experiments by varying the identity cutoffs between testing targets and training data in DUD-E. We set the identify cutoff as 90\%, 60\%, 30\% and HMM, respectively, where the sequence identity decreases progressively. According to the results shown in Figure~\ref{fig: exclusion}, we may notice that no matter under any homology exclusion threshold, S$^2$Drug can always achieve much higher virtual screening accuracy over DrugCLIP. This validates that S$^2$Drug indeed exhibits strong generalizability even under HMM scenario which enforces strict remote homology exclusion. Moreover, we observe that under 90\% and 60\% identity cutoff settings, S$^2$ still outperforms the performance of DrugHash and DrugCLIP obtained under the no exclusion setting. This evidence suggests our method significantly reduces the dependency on high train-test distribution similarity. This advantage can be attributed to the bilateral data sampling strategy used during the pretraining, which effectively mitigates bias and overfitting by filtering out redundant proteins and noisy ligands.

\begin{figure}[h]
    \centering
    \includegraphics[width=0.5\textwidth]{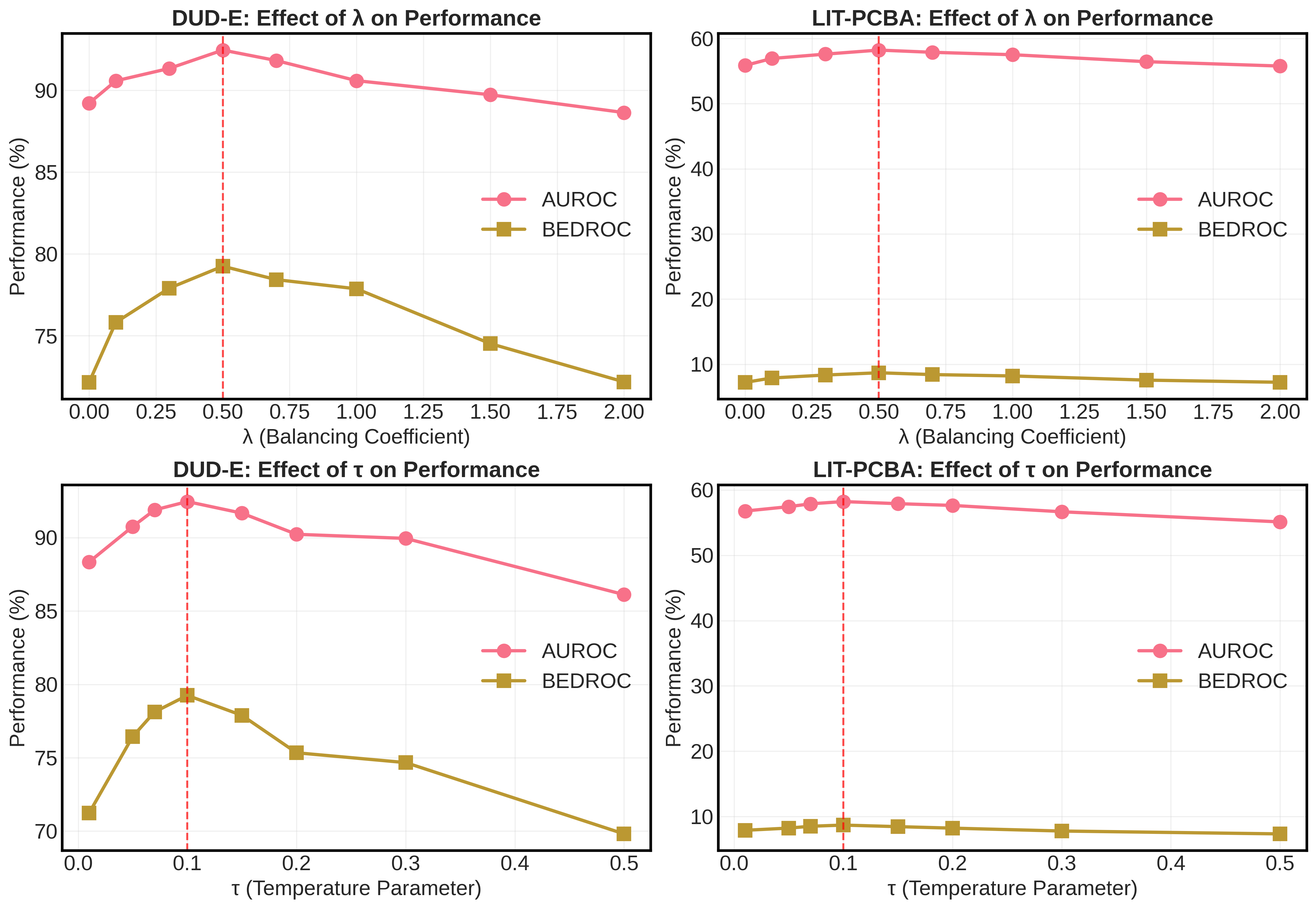}
    \caption{The hyperparameter robustness analysis for balancing coefficient $\lambda$ and temperature parameter $\tau$.}
   \label{fig: hyperparameter}
    \vspace{-3mm}
\end{figure}

\begin{table*}[h]
\centering
\resizebox{0.95\textwidth}{!}{
\begin{tabular}{l|c|c|c|c|c|c|c|c|c|c}
\toprule
\multirow{2}{*}{} & \multicolumn{5}{c|}{DUD-E} & \multicolumn{5}{c}{LIT-PCBA} \\
\cmidrule(){2-11}
& AUROC & BEDROC & EF$^{0.5\%}$ & EF$^{1\%}$ & EF$^{5\%}$ & AUROC & BEDROC & EF$^{0.5\%}$ & EF$^{1\%}$ & EF$^{5\%}$\\
\midrule
- BDS & 88.73 & 73.91 & 50.83 & 38.42 & 16.54 & 56.12 & 7.21 & 10.06 & 6.41 & 2.66 \\
- SSF & 87.92 & 69.85  & 47.14 & 35.39 & 15.12 & 55.03 & 6.88 & 9.73 & 6.15 & 2.51\\
- BSP & 89.58 & 75.60 & 52.93 & 39.27 & 17.36  &56.47  &7.56  & 10.49 & 6.85 & 2.73 \\
\midrule
S$^2$Drug & \textbf{92.46*} & \textbf{79.25*} & \textbf{58.37*} & \textbf{43.06*} & \textbf{18.82*} & \textbf{58.23*} & \textbf{8.69*} & \textbf{11.44*}  & \textbf{7.38*} & \textbf{2.97*} \\
\bottomrule
\end{tabular}}
\caption{Ablation studies on both DUD-E and LIT-PCBA datasets. The BDS, SSF, and BSP indicate bilateral data sampling, sequence-structure fusion module, and auxiliary binding site prediction task, respectively. The bold indicates the best result.}
\label{tab:ablation}
\end{table*}

\begin{table*}[h]
\centering
\resizebox{0.7\textwidth}{!}{
\begin{tabular}{lc|c|c|c|c|c}
\toprule
& & P2Rank & VN-EGNN & DiffDock & VN-EGNNrank & S$^2$Drug \\
\midrule
\multirow{2}{*}{HOLO4K} & F1 Score & 42.63 & 47.25 & \textbf{52.91} & 48.84 & 51.66 \\
& PR-AUC & 36.20 & 41.17 & \textbf{47.52}  & 44.51  & 46.97 \\
\midrule
\multirow{2}{*}{COACH420} & F1 Score & 39.14 & 40.80 & 42.25 & 43.91 & \textbf{47.32*} \\
 & PR-AUC & 34.76 & 35.95 & 38.58 & 39.81 & \textbf{42.62*} \\
\midrule
\multirow{2}{*}{ASD} & F1 Score & 33.67  & 38.93 & 40.60 & 42.18 & \textbf{48.79*} \\
 & PR-AUC & 29.58 & 32.81 & 34.41 & 37.44 & \textbf{43.96*} \\
\bottomrule
\end{tabular}}
\caption{Binding site prediction performance comparison of different methods across three datasets. Bold denotes the best.}
\vspace{-3mm}
\label{tab: binding site}
\end{table*}

\paragraph{Hyperparameter Robustness Analysis}
To investigate whether the performance of S$^2$Drug is robust to hyperparameter settings, we vary the values of the balancing coefficient $\lambda$ in the combined training objective during finetuning stage (Eq.~\ref{equ:combined}) and the temperature parameter $\tau$ in the contrastive learning objectives (Eq.~\ref{equ:pretrain contrastive} and \ref{equ:finetune contrastive}). The experiment results have provided in the curves of Figure~\ref{fig: hyperparameter}. From the table, we can first find the virtual screening accuracy is relatively robust, especially within the area around the default hyperparameter setting ($\lambda$ = 0.50, $\tau$ = 0.1). For $\lambda$, when its value is set higher than 1.0, this auxiliary binding site prediction task will introduce the optimization conflict and distract model learning away from the primary virtual screening task, leading to underfitting. As for extreme settings of $\tau$, overly low or high values may lead to over-sharpening and over-smoothing issues for learning, respectively.

\paragraph{Ablation Studies}
To explore the contribution of each proposed technique to the final virtual screening performance, we conduct experiments by individually removing the bilateral data sampling, the sequence-structure fusion module (only finetuning protein sequence encoder on PDBBind data), and the auxiliary binding site prediction task from the overall training framework, respectively. As shown in Table~\ref{tab:ablation}, the removal of any of these modules leads to a significant performance drop across all evaluation metrics. This means each of such three parts brings the positive contribution to the overall framework effectiveness. Notably, the lack of sequence-structural fusion module will result in 9.40 and 1.81 points BEDROC decrease on DUDE-E and LIT-PCBA, respectively, highlighting the importance of bridging sequence and structural information for virtual screening.

\paragraph{Auxiliary Binding Site Prediction Performance}
To validate that our S$^2$Drug framework benefits both virtual screening and binding site prediction tasks simultaneously, we empirically compare it against several representative binding site prediction baselines. According to the results in Table~\ref{tab: binding site}, we can observe that S$^2$Drug achieves the highest accuracy on 2 out of 3 benchmarks-COACH240 and ASD-showing clear superiority over the strongest baseline. 
This also validates the central hypothesis of our framework: the binding site prediction task and virtual screening task enhance each other. Binding site information provides useful structural priors for virtual screening, and vice versa. On the evaluation dataset HOLO4K, although S$^2$Drug ranks the second, it still achieves a competitive performance with a 51.66\% F1 Score and 46.97\% PR-AUC, which are closed enough to the baseline DiffDock. 
This narrow gap can be attributed to DiffDock's diffusion-based generative nature, which excels at handling symmetric complexes (common in HOLO4K) by sampling multi-modal pose distributions and mitigating geometric redundancy from symmetric units; these properties ultimately contribute to its stronger results.


\section{Conclusions and Future Works}
In this paper, we propose S$^2$Drug, a two-stage contrastive representation learning framework that explicitly bridges protein sequence and 3D structural information to enhance virtual screening. 
Our approach pretrains a sequence encoder on large-scale protein–ligand data with bilateral sampling to reduce noise, and fine-tunes by fusing sequence and 3D pocket representations with an auxiliary binding site prediction task for improved spatial specificity.
Extensive experiments across diverse benchmarks demonstrate that S$^2$Drug consistently outperforms existing structure-only and sequence-only baselines on VS accuracy. Additionally, our model achieves accurate binding site localization, further validating the complementarity between the two tasks.
Future work will explore the integration of protein surface and solvent features to enhance VS accuracy.

\section{Acknowledgments}
This work is supported by National Key R\&D Program of China No.2025ZD1803103, Beijing Academy of Artificial Intelligence. Besides, this work is supported by the Early Career Scheme (No. CityU 21219323) and the General Research Fund (No. CityU 11220324) of the University Grants Committee (UGC), the NSFC Young Scientists Fund (No. 9240127), and the Donation for Research Projects (No. 9229164 and No. 9220187).

\bibliography{aaai2026}

\begin{thebibliography}{41}
\providecommand{\natexlab}[1]{#1}

\bibitem[{Abramson et~al.(2024)Abramson, Adler, Dunger, Evans, Green, Pritzel, Ronneberger, Willmore, Ballard, Bambrick et~al.}]{abramson2024accurate}
Abramson, J.; Adler, J.; Dunger, J.; Evans, R.; Green, T.; Pritzel, A.; Ronneberger, O.; Willmore, L.; Ballard, A.~J.; Bambrick, J.; et~al. 2024.
\newblock Accurate structure prediction of biomolecular interactions with AlphaFold 3.
\newblock \emph{Nature}, 630(8016): 493--500.

\bibitem[{Alley et~al.(2019)Alley, Khimulya, Biswas, AlQuraishi, and Church}]{alley2019unified}
Alley, E.~C.; Khimulya, G.; Biswas, S.; AlQuraishi, M.; and Church, G.~M. 2019.
\newblock Unified rational protein engineering with sequence-based deep representation learning.
\newblock \emph{Nature methods}, 16(12): 1315--1322.

\bibitem[{Anfinsen(1973)}]{anfinsen1973principles}
Anfinsen, C.~B. 1973.
\newblock Principles that govern the folding of protein chains.
\newblock \emph{Science}, 181(4096): 223--230.

\bibitem[{Ballester and Mitchell(2010)}]{ballester2010machine}
Ballester, P.~J.; and Mitchell, J.~B. 2010.
\newblock A machine learning approach to predicting protein--ligand binding affinity with applications to molecular docking.
\newblock \emph{Bioinformatics}, 26(9): 1169--1175.

\bibitem[{Brocidiacono et~al.(2023)Brocidiacono, Francoeur, Aggarwal, Popov, Koes, and Tropsha}]{brocidiacono2023bigbind}
Brocidiacono, M.; Francoeur, P.; Aggarwal, R.; Popov, K.~I.; Koes, D.~R.; and Tropsha, A. 2023.
\newblock BigBind: learning from nonstructural data for structure-based virtual screening.
\newblock \emph{Journal of Chemical Information and Modeling}, 64(7): 2488--2495.

\bibitem[{Corso et~al.(2023)Corso, St{\~A}, Jing, Barzilay, Jaakkola et~al.}]{corso2023diffdock}
Corso, G.; St{\~A}, H.; Jing, B.; Barzilay, R.; Jaakkola, T.; et~al. 2023.
\newblock DiffDock: Diffusion Steps, Twists, and Turns for Molecular Docking.
\newblock In \emph{International Conference on Learning Representations (ICLR 2023)}.

\bibitem[{Durrant and McCammon(2011)}]{durrant2011nnscore}
Durrant, J.~D.; and McCammon, J.~A. 2011.
\newblock NNScore 2.0: a neural-network receptor--ligand scoring function.
\newblock \emph{Journal of chemical information and modeling}, 51(11): 2897--2903.

\bibitem[{Friesner et~al.(2004)Friesner, Banks, Murphy, Halgren, Klicic, Mainz, Repasky, Knoll, Shelley, Perry et~al.}]{friesner2004glide}
Friesner, R.~A.; Banks, J.~L.; Murphy, R.~B.; Halgren, T.~A.; Klicic, J.~J.; Mainz, D.~T.; Repasky, M.~P.; Knoll, E.~H.; Shelley, M.; Perry, J.~K.; et~al. 2004.
\newblock Glide: a new approach for rapid, accurate docking and scoring. 1. Method and assessment of docking accuracy.
\newblock \emph{Journal of medicinal chemistry}, 47(7): 1739--1749.

\bibitem[{Gao et~al.(2023)Gao, Qiang, Tan, Jia, Ren, Lu, Liu, Ma, and Lan}]{gao2023drugclip}
Gao, B.; Qiang, B.; Tan, H.; Jia, Y.; Ren, M.; Lu, M.; Liu, J.; Ma, W.-Y.; and Lan, Y. 2023.
\newblock Drugclip: Contrastive protein-molecule representation learning for virtual screening.
\newblock \emph{Advances in Neural Information Processing Systems}, 36: 44595--44614.

\bibitem[{Gaulton et~al.(2012)Gaulton, Bellis, Bento, Chambers, Davies, Hersey, Light, McGlinchey, Michalovich, Al-Lazikani et~al.}]{gaulton2012chembl}
Gaulton, A.; Bellis, L.~J.; Bento, A.~P.; Chambers, J.; Davies, M.; Hersey, A.; Light, Y.; McGlinchey, S.; Michalovich, D.; Al-Lazikani, B.; et~al. 2012.
\newblock ChEMBL: a large-scale bioactivity database for drug discovery.
\newblock \emph{Nucleic acids research}, 40(D1): D1100--D1107.

\bibitem[{Han, Hong, and Li(2025)}]{han2025drughash}
Han, J.; Hong, Y.; and Li, W.-J. 2025.
\newblock DrugHash: Hashing Based Contrastive Learning for Virtual Screening.
\newblock In \emph{Proceedings of the AAAI Conference on Artificial Intelligence}, volume~39, 17041--17049.

\bibitem[{Hu et~al.(2024)Hu, Tan, Xia, Liu, Wu, Zheng, Xu, Huang, and Li}]{hu2024learning}
Hu, B.; Tan, C.; Xia, J.; Liu, Y.; Wu, L.; Zheng, J.; Xu, Y.; Huang, Y.; and Li, S.~Z. 2024.
\newblock Learning complete protein representation by dynamically coupling of sequence and structure.
\newblock \emph{Advances in Neural Information Processing Systems}, 37: 137673--137697.

\bibitem[{Irwin and Shoichet(2005)}]{irwin2005zinc}
Irwin, J.~J.; and Shoichet, B.~K. 2005.
\newblock ZINC- a free database of commercially available compounds for virtual screening.
\newblock \emph{Journal of chemical information and modeling}, 45(1): 177--182.

\bibitem[{Kallenborn et~al.(2024)Kallenborn, Chacon, Hundt, Sirelkhatim, Didi, Cha, Dallago, Mirdita, Schmidt, and Steinegger}]{kallenborn2024gpu}
Kallenborn, F.; Chacon, A.; Hundt, C.; Sirelkhatim, H.; Didi, K.; Cha, S.; Dallago, C.; Mirdita, M.; Schmidt, B.; and Steinegger, M. 2024.
\newblock GPU-accelerated homology search with MMseqs2.
\newblock \emph{bioRxiv}, 2024--11.

\bibitem[{Kriv{\'a}k and Hoksza(2018)}]{krivak2018p2rank}
Kriv{\'a}k, R.; and Hoksza, D. 2018.
\newblock P2Rank: machine learning based tool for rapid and accurate prediction of ligand binding sites from protein structure.
\newblock \emph{Journal of cheminformatics}, 10: 1--12.

\bibitem[{Lee et~al.(2023)Lee, Yu, Lee, and Kim}]{lee2023pre}
Lee, Y.; Yu, H.; Lee, J.; and Kim, J. 2023.
\newblock Pre-training sequence, structure, and surface features for comprehensive protein representation learning.
\newblock In \emph{The Twelfth International Conference on Learning Representations}.

\bibitem[{Li et~al.(2024)Li, Li, Zhu, and Zhang}]{li2024improving}
Li, Z.; Li, M.; Zhu, L.; and Zhang, W. 2024.
\newblock Improving PTM site prediction by coupling of multi-granularity structure and multi-scale sequence representation.
\newblock In \emph{Proceedings of the AAAI Conference on Artificial Intelligence}, volume~38, 188--196.

\bibitem[{Lin et~al.(2023)Lin, Akin, Rao, Hie, Zhu, Lu, Smetanin, Verkuil, Kabeli, Shmueli et~al.}]{lin2023evolutionary}
Lin, Z.; Akin, H.; Rao, R.; Hie, B.; Zhu, Z.; Lu, W.; Smetanin, N.; Verkuil, R.; Kabeli, O.; Shmueli, Y.; et~al. 2023.
\newblock Evolutionary-scale prediction of atomic-level protein structure with a language model.
\newblock \emph{Science}, 379(6637): 1123--1130.

\bibitem[{Liu et~al.(2020)Liu, Lu, Song, Shen, Ni, Li, He, Zhang, Wang, Chen et~al.}]{liu2020unraveling}
Liu, X.; Lu, S.; Song, K.; Shen, Q.; Ni, D.; Li, Q.; He, X.; Zhang, H.; Wang, Q.; Chen, Y.; et~al. 2020.
\newblock Unraveling allosteric landscapes of allosterome with ASD.
\newblock \emph{Nucleic acids research}, 48(D1): D394--D401.

\bibitem[{Lyu, Irwin, and Shoichet(2023)}]{lyu2023modeling}
Lyu, J.; Irwin, J.~J.; and Shoichet, B.~K. 2023.
\newblock Modeling the expansion of virtual screening libraries.
\newblock \emph{Nature chemical biology}, 19(6): 712--718.

\bibitem[{Maia et~al.(2020)Maia, Assis, De~Oliveira, Da~Silva, and Taranto}]{maia2020structure}
Maia, E. H.~B.; Assis, L.~C.; De~Oliveira, T.~A.; Da~Silva, A.~M.; and Taranto, A.~G. 2020.
\newblock Structure-based virtual screening: from classical to artificial intelligence.
\newblock \emph{Frontiers in chemistry}, 8: 343.

\bibitem[{McNutt et~al.(2021)McNutt, Francoeur, Aggarwal, Masuda, Meli, Ragoza, Sunseri, and Koes}]{mcnutt2021gnina}
McNutt, A.~T.; Francoeur, P.; Aggarwal, R.; Masuda, T.; Meli, R.; Ragoza, M.; Sunseri, J.; and Koes, D.~R. 2021.
\newblock GNINA 1.0: molecular docking with deep learning.
\newblock \emph{Journal of cheminformatics}, 13(1): 43.

\bibitem[{Mysinger et~al.(2012)Mysinger, Carchia, Irwin, and Shoichet}]{mysinger2012directory}
Mysinger, M.~M.; Carchia, M.; Irwin, J.~J.; and Shoichet, B.~K. 2012.
\newblock Directory of useful decoys, enhanced (DUD-E): better ligands and decoys for better benchmarking.
\newblock \emph{Journal of medicinal chemistry}, 55(14): 6582--6594.

\bibitem[{Oord(2018)}]{oord2018representation}
Oord, A. 2018.
\newblock Representation learning with contrastive predictive coding.
\newblock \emph{arXiv preprint arXiv:1807.03748}.

\bibitem[{{\"O}zt{\"u}rk, {\"O}zg{\"u}r, and Ozkirimli(2018)}]{ozturk2018deepdta}
{\"O}zt{\"u}rk, H.; {\"O}zg{\"u}r, A.; and Ozkirimli, E. 2018.
\newblock DeepDTA: deep drug--target binding affinity prediction.
\newblock \emph{Bioinformatics}, 34(17): i821--i829.

\bibitem[{Rives et~al.(2021)Rives, Meier, Sercu, Goyal, Lin, Liu, Guo, Ott, Zitnick, Ma et~al.}]{rives2021biological}
Rives, A.; Meier, J.; Sercu, T.; Goyal, S.; Lin, Z.; Liu, J.; Guo, D.; Ott, M.; Zitnick, C.~L.; Ma, J.; et~al. 2021.
\newblock Biological structure and function emerge from scaling unsupervised learning to 250 million protein sequences.
\newblock \emph{Proceedings of the National Academy of Sciences}, 118(15): e2016239118.

\bibitem[{Schneckenreiter et~al.(2025)Schneckenreiter, Luukkonen, Friedrich, Kuhn, and Klambauer}]{schneckenreiter2025ligand}
Schneckenreiter, L.; Luukkonen, S.; Friedrich, L.; Kuhn, D.; and Klambauer, G. 2025.
\newblock Ligand-Conditioned Binding Site Prediction Using Contrastive Geometric Learning.
\newblock In \emph{Learning Meaningful Representations of Life (LMRL) Workshop at ICLR 2025}.

\bibitem[{Sestak et~al.(2023)Sestak, Schneckenreiter, Hochreiter, Mayr, and Klambauer}]{sestak2023vn}
Sestak, F.; Schneckenreiter, L.; Hochreiter, S.; Mayr, A.; and Klambauer, G. 2023.
\newblock VN-EGNN: Equivariant Graph Neural Networks with Virtual Nodes Enhance Protein Binding Site Identification.
\newblock In \emph{NeurIPS 2023 Workshop: New Frontiers in Graph Learning}.

\bibitem[{Shen et~al.(2023)Shen, Feng, Qiu, and Wei}]{shen2023svsbi}
Shen, L.; Feng, H.; Qiu, Y.; and Wei, G.-W. 2023.
\newblock SVSBI: sequence-based virtual screening of biomolecular interactions.
\newblock \emph{Communications biology}, 6(1): 536.

\bibitem[{Shivanyuk et~al.(2007)Shivanyuk, Ryabukhin, Tolmachev, Bogolyubsky, Mykytenko, Chupryna, Heilman, and Kostyuk}]{shivanyuk2007enamine}
Shivanyuk, A.~N.; Ryabukhin, S.~V.; Tolmachev, A.; Bogolyubsky, A.; Mykytenko, D.; Chupryna, A.; Heilman, W.; and Kostyuk, A. 2007.
\newblock Enamine real database: Making chemical diversity real.
\newblock \emph{Chemistry today}, 25(6): 58--59.

\bibitem[{Shoichet(2004)}]{shoichet2004virtual}
Shoichet, B.~K. 2004.
\newblock Virtual screening of chemical libraries.
\newblock \emph{Nature}, 432(7019): 862--865.

\bibitem[{Spitzer and Jain(2012)}]{spitzer2012surflex}
Spitzer, R.; and Jain, A.~N. 2012.
\newblock Surflex-Dock: Docking benchmarks and real-world application.
\newblock \emph{Journal of computer-aided molecular design}, 26: 687--699.

\bibitem[{Stepniewska-Dziubinska, Zielenkiewicz, and Siedlecki(2017)}]{stepniewska2017pafnucy}
Stepniewska-Dziubinska, M.~M.; Zielenkiewicz, P.; and Siedlecki, P. 2017.
\newblock Pafnucy-a deep neural network for structure-based drug discovery.
\newblock \emph{Bioinformatics}, 1050: 19.

\bibitem[{Tran-Nguyen, Jacquemard, and Rognan(2020)}]{tran2020lit}
Tran-Nguyen, V.-K.; Jacquemard, C.; and Rognan, D. 2020.
\newblock LIT-PCBA: an unbiased data set for machine learning and virtual screening.
\newblock \emph{Journal of chemical information and modeling}, 60(9): 4263--4273.

\bibitem[{Trott and Olson(2010)}]{trott2010autodock}
Trott, O.; and Olson, A.~J. 2010.
\newblock AutoDock Vina: improving the speed and accuracy of docking with a new scoring function, efficient optimization, and multithreading.
\newblock \emph{Journal of computational chemistry}, 31(2): 455--461.

\bibitem[{Wang et~al.(2022)Wang, Wang, Cao, and Barati~Farimani}]{wang2022molecular}
Wang, Y.; Wang, J.; Cao, Z.; and Barati~Farimani, A. 2022.
\newblock Molecular contrastive learning of representations via graph neural networks.
\newblock \emph{Nature Machine Intelligence}, 4(3): 279--287.

\bibitem[{Wu et~al.(2022)Wu, Gao, Zeng, Zhang, and Li}]{wu2022bridgedpi}
Wu, Y.; Gao, M.; Zeng, M.; Zhang, J.; and Li, M. 2022.
\newblock BridgeDPI: a novel graph neural network for predicting drug--protein interactions.
\newblock \emph{Bioinformatics}, 38(9): 2571--2578.

\bibitem[{Yazdani-Jahromi et~al.(2022)Yazdani-Jahromi, Yousefi, Tayebi, Kolanthai, Neal, Seal, and Garibay}]{yazdani2022attentionsitedti}
Yazdani-Jahromi, M.; Yousefi, N.; Tayebi, A.; Kolanthai, E.; Neal, C.~J.; Seal, S.; and Garibay, O.~O. 2022.
\newblock AttentionSiteDTI: an interpretable graph-based model for drug-target interaction prediction using NLP sentence-level relation classification.
\newblock \emph{Briefings in Bioinformatics}, 23(4): bbac272.

\bibitem[{Zhang et~al.(2023)Zhang, Gao, Wang, Chen, Zhang, Chen, Li, Qi, and Wang}]{zhang2023planet}
Zhang, X.; Gao, H.; Wang, H.; Chen, Z.; Zhang, Z.; Chen, X.; Li, Y.; Qi, Y.; and Wang, R. 2023.
\newblock Planet: a multi-objective graph neural network model for protein--ligand binding affinity prediction.
\newblock \emph{Journal of Chemical Information and Modeling}, 64(7): 2205--2220.

\bibitem[{Zheng, Fan, and Mu(2019)}]{zheng2019onionnet}
Zheng, L.; Fan, J.; and Mu, Y. 2019.
\newblock Onionnet: a multiple-layer intermolecular-contact-based convolutional neural network for protein--ligand binding affinity prediction.
\newblock \emph{ACS omega}, 4(14): 15956--15965.

\bibitem[{Zhou et~al.(2023)Zhou, Gao, Ding, Zheng, Xu, Wei, Zhang, and Ke}]{zhouuni}
Zhou, G.; Gao, Z.; Ding, Q.; Zheng, H.; Xu, H.; Wei, Z.; Zhang, L.; and Ke, G. 2023.
\newblock Uni-Mol: A Universal 3D Molecular Representation Learning Framework.
\newblock In \emph{The Eleventh International Conference on Learning Representations}.

\end{thebibliography}


\section{Detailed Datasets Introduction}
The evaluation of S$^2$Drug involves two main categories of tasks: virtual screening and binding site prediction. Each task employs specific benchmark datasets that are widely recognized in the computational drug discovery community.

\subsection{Virtual Screening Datasets}
Following previous works in this area~\citep{gao2023drugclip, han2025drughash}, we conduct the virtual screening task on two most representative datasets as follows:
\begin{itemize}[leftmargin=*, topsep=2pt]
    \item \textbf{DUD-E}\citep{mysinger2012directory}: DUD-E is one of the most widely used benchmarks for evaluating virtual screening methods. Originally introduced by Mysinger et al. (2012), DUD-E addresses several limitations of earlier virtual screening datasets by providing improved ligand and decoy selection criteria. The dataset contains carefully curated protein targets, each paired with known active ligands and computationally generated decoy molecules. The decoy selection strategy in DUD-E is particularly sophisticated, where decoys are selected to be chemically similar to active ligands in terms of molecular properties such as molecular weight, logP, rotatable bonds, and hydrogen bond donors/acceptors, but are assumed to be inactive based on topological dissimilarity. The enhanced version addresses potential biases present in the original DUD dataset, such as artificial enrichment due to trivial chemical differences between actives and decoys. DUD-E covers a wide range of protein families and binding sites, making it suitable for assessing method generalizability across diverse biological targets. The evaluation protocol follows a zero-shot setting where methods are tested on targets not seen during training, ensuring robust assessment of true predictive capability.
    \item \textbf{LIT-PCBA}~\citep{tran2020lit}: LIT-PCBA represents a more challenging and realistic virtual screening benchmark that was specifically designed to address limitations in existing benchmarks and provide a more unbiased evaluation framework. Unlike many datasets that rely on high-throughput screening data, LIT-PCBA extracts active compounds from peer-reviewed literature, potentially representing higher-quality and more relevant binding interactions that have been subjected to rigorous scientific scrutiny. The dataset maintains the natural distribution of active versus inactive compounds found in real drug discovery campaigns, typically characterized by very low hit rates that reflect the genuine difficulty of identifying bioactive molecules. LIT-PCBA has been carefully curated to minimize artificial patterns that could lead to inflated performance estimates, addressing a common criticism of earlier benchmarks where methods could achieve high performance by exploiting dataset-specific biases rather than learning genuine structure-activity relationships. The dataset encompasses various assay formats and biological endpoints, reflecting the diversity of real-world drug discovery efforts and providing a more comprehensive assessment of virtual screening methods across different experimental contexts.
\end{itemize}

\subsection{Binding Site Prediction Datasets}
The auxiliary binding site prediction task is evaluated on three established datasets, each offering unique challenges and characteristics.

\begin{itemize}[leftmargin=*, topsep=2pt]
    \item \textbf{HOLO4K}~\citep{krivak2018p2rank}: HOLO4K is a comprehensive binding site prediction benchmark that focuses on holo protein structures (proteins in complex with ligands). This dataset is particularly valuable for evaluating methods' ability to identify binding sites in their bound conformations.
    \item \textbf{COACH420}~\citep{krivak2018p2rank}: COACH420 is a well-established benchmark for binding site prediction methods. The dataset contains 420 protein structures with experimentally determined binding sites, providing ground truth annotations for rigorous evaluation.
    \item \textbf{ASD}~\citep{liu2020unraveling}: This dataset specifically focuses on allosteric binding sites, which represent a particularly challenging subset of binding site prediction. Allosteric sites are often distant from active sites and may exhibit more subtle structural features, making their identification more difficult than traditional orthosteric binding sites.
\end{itemize}

\section{Detailed Baselines Introduction}
The experimental evaluation compares S$^2$Drug against a comprehensive set of baseline methods representing both traditional computational approaches and state-of-the-art machine learning techniques.

\subsection{Virtual Screening Baselines}
\paragraph{Docking-based Methods}
Docking-based methods represent the classical approaches to virtual screening, simulating the physical binding process between ligands and proteins to estimate binding affinity.
\begin{itemize}[leftmargin=*, topsep=2pt]
   \item \textbf{Glide-SP}~\citep{friesner2004glide}: This is a widely-used commercial docking software that employs a hierarchical approach to ligand docking. It uses a series of increasingly sophisticated scoring functions to evaluate ligand poses, balancing accuracy with computational efficiency. Glide-SP is particularly known for its ability to handle flexible ligand conformations and its robust scoring function that considers various intermolecular interactions.
   \item \textbf{Vina}~\citep{trott2010autodock}: Vina represents a significant improvement over the original AutoDock software. It features a new scoring function, efficient optimization algorithms, and multithreading capabilities. Vina is notable for its speed improvements (up to two orders of magnitude faster than AutoDock 4) while maintaining or improving docking accuracy. The software uses a sophisticated algorithm for exploring conformational space and evaluating binding poses.
   \item \textbf{Surflex}~\citep{spitzer2012surflex}: This is another established docking program that uses a surface-based molecular similarity approach. It generates ligand poses by aligning molecular fragments to complementary regions of the protein binding site, making it particularly effective for fragment-based drug design applications.
\end{itemize}

\paragraph{Learning-based Methods}
These methods use machine learning to either (i) predict binding affinity as a continuous value (regression), (ii) treat virtual screening as a binary classification problem, or (iii) formulate it as a retrieval task, representing recent modeling paradigms for this problem.
\begin{itemize}[leftmargin=*, topsep=2pt]
  \item \textbf{NN-Score}~\citep{durrant2011nnscore}: NN-Score employs neural networks to score protein-ligand complexes. The method extracts various structural features from protein-ligand complexes and uses these as input to a neural network that predicts binding affinity. NN-Score 2.0 represents an improved version with enhanced feature selection and network architecture.
  \item \textbf{RF-Score}~\citep{ballester2010machine}:RF-Score utilizes Random Forest algorithms to predict protein-ligand binding affinity. The method extracts atom-type pair counts as features and uses ensemble learning to make predictions. RF-Score was one of the first successful applications of machine learning to binding affinity prediction and demonstrated superior performance to many traditional scoring functions.
  \item \textbf{Pafnucy}~\citep{stepniewska2017pafnucy}: Pafnucy represents an early application of deep learning to structure-based drug discovery. It uses 3D convolutional neural networks to analyze protein-ligand complexes, treating the binding site as a 3D grid and learning spatial patterns associated with binding affinity.
  \item \textbf{Planet}~\citep{zhang2023planet}: Planet is a multi-objective graph neural network model specifically designed for protein-ligand binding affinity prediction. It incorporates both local and global structural information through graph representations and aims to balance multiple objectives in the learning process.
  \item \textbf{DeepDTA}~\citep{ozturk2018deepdta}: DeepDTA focuses on drug-target binding affinity prediction using sequence information. The method employs convolutional neural networks to process both drug SMILES strings and protein sequences, learning representations that capture binding-relevant patterns.
  \item \textbf{Gnina}~\citep{mcnutt2021gnina}: Gnina represents a hybrid approach combining traditional docking with deep learning. It uses convolutional neural networks to score docked poses, providing a more accurate alternative to traditional scoring functions while maintaining the geometric insights of molecular docking.
  \item \textbf{Banana}~\citep{brocidiacono2023bigbind}: Banana focuses on learning from non-structural data for structure-based virtual screening. The method addresses the challenge of limited structural data by incorporating alternative data sources and representations.
  \item \textbf{SVSBI}~\citep{shen2023svsbi}: SVSBI represents an early attempt to incorporate protein sequence information into virtual screening. The method uses sequence-based features to predict protein-ligand interactions, addressing scenarios where structural information may be limited or unavailable.
  \item \textbf{DrugCLIP}~\citep{gao2023drugclip}: DrugCLIP applies contrastive learning principles to protein-ligand representation learning. The method uses a CLIP-like architecture to learn aligned representations of proteins and ligands, enabling efficient similarity-based virtual screening. DrugCLIP employs SE(3) Transformers to encode 3D structural information and has demonstrated strong performance on virtual screening benchmarks.
  \item \textbf{DrugHash}~\citep{han2025drughash}: DrugHash represents the most recent advancement in contrastive learning for virtual screening. The method uses hashing-based techniques to enable efficient large-scale virtual screening while maintaining high accuracy. DrugHash builds upon structural encoding approaches but introduces novel hashing mechanisms for scalable similarity search.
\end{itemize}

\subsection{Binding Site Prediction Baselines}
The binding site prediction task employs specialized methods designed to identify functional binding regions within protein structures.
\begin{itemize}[leftmargin=*, topsep=2pt]
    \item \textbf{P2Rank}~\citep{krivak2018p2rank}: P2Rank is a machine learning-based tool for rapid and accurate prediction of ligand binding sites from protein structure. The method uses Random Forest algorithms with carefully selected structural features, including geometric and physicochemical properties of protein surface regions. P2Rank has become a widely-used baseline due to its balance of accuracy and computational efficiency.
    \item \textbf{VN-EGNN}~\citep{sestak2023vn}: VN-EGNN represents a graph neural network approach to binding site prediction. The method uses equivariant neural networks that respect the geometric symmetries inherent in protein structures, potentially providing more robust predictions across different protein orientations and conformations.
    \item \textbf{DiffDock}~\citep{corso2023diffdock}: DiffDock employs diffusion models for molecular docking and binding site prediction. The method uses generative modeling approaches to sample probable binding poses and identify likely binding regions. DiffDock's diffusion-based approach allows it to handle multi-modal binding distributions and geometric uncertainties effectively.
    \item \textbf{VN-EGNNrank}~\citep{schneckenreiter2025ligand}: VN-EGNNrank extends equivariant graph neural networks (VN-EGNN) with ranking-based learning objectives. The method focuses on learning relative preferences between potential binding sites, potentially providing more robust predictions in challenging scenarios.
\end{itemize}

\section{Efficiency Analysis}
To evaluate the computational efficiency of S$^2$Drug framework, we conduct comprehensive experiments and compare with baseline methods on training time, inference speed, and memory consumption during training and inference phases.

\begin{table}[h]
\centering
\resizebox{0.48\textwidth}{!}{
\begin{tabular}{l|c|c|c}
\toprule
Method & Pretraining & Finetuning & Total Training  \\
\midrule
Planet & - & 14.2 hours & 14.2 hours  \\
Banana & - & 18.6 hours & 18.6 hours  \\
SVSBI & - & 17.1 hours & 17.1 hours  \\
DrugCLIP & - & 16.3 hours & 16.3 hours  \\
DrugHash & - & 20.8 hours &  20.8 hours \\
\midrule
S$^2$Drug & 8.5 hours  & 6.2 hours & 14.7 hours  \\
S$^2$Drug-Hash & 8.5 hours  & 8.3 hours & 16.8 hours \\
\bottomrule
\end{tabular}}
\caption{Training time comparison between S$^2$Drug and competitive learning-based virtual screening baselines.}
\label{tab:training time}
\end{table}

\subsection{Training Time Cost}
We measure the training time for both pretraining and finetuning stages of S$^2$Drug compared to other learning-based baselines. Table \ref{tab:training time} shows the training time comparison on our experimental setup (8 $\times$ NVIDIA RTX A6000 GPUs). Although S$^2$Drug requires additional pretraining time due to its two-stage design, the finetuning stage is significantly faster than baseline methods. This is because the sequence pretraining provides robust initial representations, leading to faster convergence during structure-sequence fusion finetuning, thus requiring less training epochs for fulfilling stop criteria. Notably, the DrugHash consumes more training time than DrugCLIP (20.8 vs 16.3 hours), due to the iterative training procedure between normal protein/molecule representation learning and cross-modal hashing. Besides, we also integrate the binary code hashing technique proposed by DrugHash into our S$^2$Drug framework, achieving S$^2$Drug-Hash. The pretraining stage of S$^2$Drug-Hash keeps same with S$^2$Drug, while introducing cross-modal hashing and corresponding iterative training intro finetuning stage. From Table \ref{tab:training time}, we can find finetuning stage of S$^2$Drug-Hash indeed takes more time than vanilla S$^2$Drug, similar to the trend between DrugHash and DrugCLIP. Though, the finetuning time and the total time cost of S$^2$Drug-Hash are both less than the training time of DrugHash, further demonstrating the competitiveness on training efficiency.

\subsection{Inference Speed}
We evaluate the inference speed by measuring the time required to process protein-ligand pairs during virtual screening, following the experimental setup from DrugHash~\citep{han2025drughash} paper. Similar to this previous work, we pre-compute and store the protein and ligand representations in an offline way by forwarding the raw protein and ligand data through the learned encoders and fusion modules. Thus, during the inference stage, we only need to load the pre-computed protein representation vectors and ligand representation vectors, then conducting the dot product operations between them. In detail, we study the inference speed from two perspectives: 1) single pair inference, in which we only load representation vectors corresponding to a target protein and a ligand candidate each time; 2) batch inference, in which we load representation vectors for a batch of target protein and ligand candidate pairs each time. We take DUD-E~\citep{mysinger2012directory} dataset for evaluation here.

\paragraph{Single Pair Inference}
For single protein-ligand pair processing, the averaged inference time costs are as follows: Planet: 0.052 seconds per pair; Banana: 0.055 seconds per pair; DrugCLIP: 0.053 seconds per pair; DrugHash: 0.043 seconds per pair; S$^2$Drug: 0.051 seconds per pair; S$^2$Drug-Hash: 0.042 seconds per pair. From these results, we can observe vanilla S$^2$Drug can achieve similar inference efficiency to DrugCLIP and much better virtual screening accuracy, as mentioned in the main text. Meanwhile, S$^2$Drug-Hash achieve even a little bit faster inference effect compared with DrugHash. There results show our proposed S$^2$Drug and its efficient variant, S$^2$Drug-Hash, exhibit strong competitiveness on inference speed.

\begin{table}[h]
\centering
\resizebox{0.48\textwidth}{!}{
\begin{tabular}{l|c|c|c|c}
\toprule
Batch Size & DrugCLIP & DrugHash & S$^2$Drug & S$^2$Drug-Hash\\
\midrule
1 & 53.1 & 43.2 & 51.4& 42.0 \\
16 & 18.9 & 12.8 & 19.2 & 11.3 \\
64 & 12.4 &  8.1& 11.8 & 7.8\\
128 & 10.8 & 6.9 & 10.6 & 6.6\\
\bottomrule
\end{tabular}}
\caption{Inference speed comparison across different batch sizes. The unit for the values in the table is ms/pair.}
\label{tab:inference time}
\end{table}

\paragraph{Batch Inference}
For evaluating the batch inference efficiency, we vary the batch size from 1 to 128 and record the averaged inference speed per pair. The corresponding results have been provided in Table~\ref{tab:inference time}. From this table, increasing batch size indeed helps improve overall inference efficiency, though the batch size value is bounded by the GPU memory. Comparing DrugCLIP with S$^2$Drug, S$^2$Drug even achieves higher inference speed, though this kind of improvement is marginal, because they adopt the same representation vector dimension. Meanwhile, considering the virtual screening accuracy advantages of our S$^2$Drug, it is still encouraging that we do not need to sacrifice online inference efficiency for higher accuracy. Besides, though DrugHash indeed decreases the time cost during batch inference compared with DrugCLIP, our S$^2$Drug-Hash can also achieve this superiority, even faster than DrugHash by introducing the same binary hash code technique for efficient retrieval (averaged 0.83 ms/pair reduction among such batch sizes). We attribute this part of the speed improvement to the usage of sparse matrix multiplication optimization packages which are missing in baselines' original implementation. These results further validate that our proposed two-stage sequence-structure fusion learning frameworks is actually orthogonal to the hashing strategy in DrugHash. In fact, we can even integrate other techniques like model pruning and quantization to further improve the efficiency of S$^2$Drug.

\paragraph{Comparison with Docking Methods} In addition to above comparison with learning-based methods, we also take docking methods into consideration here. While docking-based methods like Glide-SP~\citep{friesner2004glide} and Vina~\citep{trott2010autodock} require 30-180 seconds per protein-ligand pair depending on search space and conformational sampling, S$^2$Drug achieves comparable or superior accuracy (as shown in the main text) in less than 0.05 seconds per pair for inference, representing a $\mathbf{600-3600 \times}$ \textbf{speedup} over traditional docking approaches.

\subsection{Memory Consumption}
We analyze peak GPU memory usage during training and inference phases, with particular focus on large-scale virtual screening scenarios.

\paragraph{Training Memory Usage}
We provide the GPU memory usage of our S$^2$Drug as follows: sequence pretraining stage: 31.2 GB (per GPU, batch size = 128); Structure-Sequence Finetuning: 35.8 GB (per GPU, batch size = 48). As for S$^2$Drug-Hash, considering its pretraining stage is exactly same as vanilla S$^2$Drug, we provide its fnetuning phase peak memory usage: 36.2 GB (per GPU, batch size = 48). Besides, we show the training memory usage of two previous contrastive learning-based virtual screening methods: DrugCLIP: 28.6 GB (per GPU, batch size = 48)
DrugHash: 29.1 GB (per GPU, batch size = 48). From these results, we can find that the memory cost of sequence pretraining is relatively low, only 31.2GB at batch size 128 compared with 28.6GB at batch size 28 of DrugCLIP. This is because we can actually freeze the parameters of ESM2 backbone and only train a lightweight adapter matrix to map ESM2 output embeddings into the protein sequence-ligand interaction representation space. The model parameter of ESM2 backbone can been only loaded to CPU rather than GPU. As for the finetuning stage, S$^2$Drug and S$^2$Drug-Hash indeed consume relatively more GPU memory than DrugCLIP and DrugHash. This is because they integrate the residue sequence encoder which is also involved into the finetuning process. But such kind of additional memory costs (6.7 GB and 7.1 GB) compared with DrugHash are still acceptable, considering the overall 35.8 GB and 36.2 GB are still within the 48GB memory capacity of each NVIDIA A6000 GPU.

\begin{table}[h]
\centering
\resizebox{0.48\textwidth}{!}{
\begin{tabular}{l|c|c|c}
\toprule
Method & DUD-E($\sim$1.2M) & ZINC ($\sim$2.3M) & REAL ($\sim$6.75B)\\
\midrule
Planet & 1.36GB & 2.57GB & 7264GB \\
Banana & 0.54GB & 1.00GB & 3080GB \\
DrugCLIP & 0.54GB & 1.00GB & 3080GB \\
DrugHash & 0.018GB &  0.03GB & 96GB \\
\midrule
S$^2$Drug & 0.54GB & 1.00GB & 3080GB \\
S$^2$Drug-Hash & 0.018GB & 0.03GB  & 96GB \\
\bottomrule
\end{tabular}}
\caption{Memory cost comparison on various virtual screening benchmarks.}
\label{tab:inference memory}
\end{table}

\paragraph{Inference Memory Usage}
To examine the memory usage during inference phase, especially in practical virtual screening scenarios, we utilize large-scale datasets including ZINC (2.3M ligand candidates)~\citep{irwin2005zinc} and Enamine REAL (6.75B candidates)~\citep{shivanyuk2007enamine}, in addition to the DUD-E(1.2M ligand candidates)~\citep{mysinger2012directory}. We don't provide the results on LIT-PCBA~\citep{tran2020lit}, considering its candidate pool size is even less than 1M, which can hardly distinguish the efficiency of different methods. Especially, we follow the experiment setting in DrugHash~\citep{han2025drughash} and load a target protein as well as all ligand candidates each time, to further ensure comparison consistency. The corresponding results have been provided in Table~\ref{tab:inference memory}. From the shown results among three datasets, we can observe that the memory usage of S$^2$Drug is the same as DrugCLIP, while the same for DrugHash and S$^2$Drug-Hash. This is because the online inference phase only involves the dot production operation, thus only loading pre-computed protein and ligand representation vectors to GPUs. As the expansion of dataset scale, especially the ligand candidate numbers, the memory usage of DrugCLIP and vanilla S$^2$Drug increases a lot  under one glance setting (3080 GB on REAL) which is definitely unacceptable on most single-node servers. Thus, the batch processing can be a alternative option. Though insisting on one glance setting, our S$^2$Drug-Hash can still provide very competitive 96 GB usage on REAL. In a word, the inference memory usage of S$^2$Drug can keep consistent with the strong and representative baseline: DrugCLIP. Even under extremely large dataset, we can also switch to its efficient variant S$^2$Drug-Hash or batch processing solution.

\subsection{Discussion}
The efficiency analysis reveals that while S$^2$Drug requires additional computational investment during the pretraining stage, it delivers substantial benefits:
\begin{itemize}[leftmargin=*, topsep=2pt]
    \item \textbf{Faster Convergence}: Sequence pretraining enables rapid convergence in finetuning stage by providing robust initial representations, reducing total training time.
    \item \textbf{Competitive Inference Speed}: Despite the sequence-structure fusion architecture, inference process remains fast enough for real-time screening applications due to only dot product operation during online inference.
    \item \textbf{Memory Efficiency}: The adapter design in sequence encoder allows larger batch size in pretraining stage, while the inference memory usage is almost same as baselines.
    \item \textbf{Superior Speed-Accuracy Trade-off}: S$^2$Drug achieves the best virtual screening performance while maintaining inference speeds orders of magnitude faster than docking-based methods. Meanwhile, its efficient variant S$^2$Drug-Hash even achieves slightly faster inference speed than fastest baseline, DrugHash.
\end{itemize}
These results demonstrate that S$^2$Drug provides an excellent balance between computational efficiency and prediction accuracy, making it practical for large-scale drug discovery applications.

\section{Limitation Analysis}
Despite the strong empirical performance and demonstrated generalizability of S$^2$Drug, several limitations remain that warrant future exploration:

\paragraph{Dependence on High-Quality Structural Data in Finetuning}
While S$^2$Drug leverages sequence data to enhance generalization, the finetuning stage still relies on precise 3D structural information from PDBBind. This restricts applicability to proteins with experimentally or computationally resolved pocket structures. In real-world scenarios where structural data is sparse, performance may degrade due to potential inaccuracies in modeled conformations.

\paragraph{Sequence-Structure Integration at Residue Level Only}
The fusion mechanism in S$^2$Drug operates at the residue level, treating the residue as a minimal structure unit. However, key interactions such as hydrogen bonding and $\pi$–$\pi$ stacking often involve finer atomic-level granularity. The current design may thus underutilize detailed atomic features that could further refine virtual screening performance.

\paragraph{Computational Overhead in Dual-Modality Learning} 
The two-stage framework, particularly the fusion of high-dimensional sequence and structural features along with the auxiliary binding site prediction task, introduces additional computational complexity. This may pose challenges for training in resource-constrained or large-scale dataset settings, where efficiency is critical.

\paragraph{Dataset and Label Bias in Pretraining and Evaluation}
Despite bilateral data sampling strategies, the ChemBL-based pretraining dataset may still contain residual bias in terms of protein family distributions and ligand scaffold coverage. Similarly, benchmark datasets like DUD-E and LIT-PCBA are known to contain decoy selection artifacts, which could potentially overestimate model performance.

\paragraph{Limited Exploration of Surface and Solvent Effects}
S$^2$Drug currently does not model solvent accessibility, electrostatics, or protein surface features, which may also related to ligand binding. These aspects may be especially important for capturing entropic and enthalpic effects in real binding.

Future research may address these limitations by incorporating predicted or coarse-grained pocket structures, introducing finer-grained interaction modeling, improving computational efficiency, and expanding toward richer biochemical contexts including surface topology and dynamic solvent interactions.

\section{Pseudocode}
To facilitate the reproduction of our proposed S$^2$Drug framework, we provide the pseudocode in Algorithm\ref{alg:S$^2$Drug}. Our framework is generaly split into two parts: sequence model pretraining and sequence-structure fusion finetuning, whose step-by-step details have also been provided.

\begin{algorithm*}[h]
\caption{Bridging Protein Sequence and 3D Structure in Contrastive Representation Learning for Virtual Screening}
\label{alg:S$^2$Drug}
\textbf{Input:} ChemBL dataset $\mathcal{D}_0$, PDBBind dataset $\mathcal{D}_{PDB}$\\
\textbf{Output:} Trained protein-ligand interaction model $\theta$
\begin{algorithmic}[1]
\State \textbf{// Stage 1: Sequence Model Pretraining}
\State \textbf{Bilateral Data Sampling:}
\State \textbf{Protein-side:} Apply homology-aware downweighting: $\Pr(P_n) = \frac{1}{|C_m^{hom}|^{0.5}}$
\State Apply functional deduplication (retain one representative per function)
\State \textbf{Ligand-side:} Apply affinity variability filtering ($\sigma_n < 1.0$)
\State Remove frequent hitters with $f(L_n) > 20$
\State \textbf{Joint Subsampling with Rebalancing:} $\mathcal{D} = \text{Sample}[\Pr(P) \cdot I_{\text{clean}}(P,L,a) \cdot w_{\text{lig}}(L)]$, where $w_{\text{lig}}(L) \propto 1/f(L)$
\State \textbf{Contrastive Learning:}
\State Initialize $\text{Seq}^p(\cdot)$ from ESM2, $\text{Stru}^l(\cdot)$ from Uni-Mol
\For{each batch $\{(P_n, L_n)\}_{n=1}^N$ from $\mathcal{D}$}
    \State $h_n^{p,s} = \text{MeanPool}(\text{Seq}^p(S(P_n)))$, $h_n^{l,g} = \text{MeanPool}(\text{Stru}^l(G(L_n)))$
    \State Compute symmetric contrastive loss $\mathcal{L}_{pc}$ (Eq. 8)
    \State Update parameters: $\theta \leftarrow \theta - \eta \nabla_\theta \mathcal{L}_{pc}(\theta)$
\EndFor
\State \textbf{// Stage 2: Sequence-Structure Fusion Finetuning}
\For{each batch $\{(P_n, L_n)\}_{n=1}^N$ from $\mathcal{D}_{PDB}$}
    \State \textbf{Sequence-Structure Fusion Module:}
    \State Get $\{x_{n,i}^s\} = \text{Seq}^p(S(P_n))$, $\{z_a\} = \text{Stru}^p(B(P_n))$
    \For{each pocket residue $r_i$}
        \State $x_{n,i}^g = \frac{1}{|r_i|} \sum_{a \in r_i} z_a$ \Comment{Atom-to-residue aggregation}
        \State $\beta_{n,i} = \sigma(W_\beta^T [W_s x_{n,i}^s; W_g x_{n,i}^g] + b_\beta)$ \Comment{Gating}
        \State $x_{n,i}^f = \beta_{n,i} \cdot W_s x_{n,i}^s + (1 - \beta_{n,i}) \cdot W_g x_{n,i}^g$
    \EndFor
    \State $h_n^{p,f} = \text{MeanPool}(\{x_{n,i}^f\})$, $h_n^{l,g} = \text{MeanPool}(\text{Stru}^l(G(L_n)))$
    \State \textbf{Binding Site Prediction:}
    \State Sample ligand probes $\{L_k\}_{k=1}^K$
    \State Compute attention: $\alpha_{n,i}^k = \text{softmax}(W_r x_{n,i}^s \cdot W_l h_k^l)$
    \State $\hat{y}_{n,i} = \frac{1}{K} \sum_{k=1}^K \alpha_{n,i}^k$
    \State \textbf{Joint Training:}
    \State $\mathcal{L}_{total} = \mathcal{L}_{fc} + \lambda \cdot \mathcal{L}_{bsp}$ where $\lambda = 0.5$
    \State Update parameters: $\theta \leftarrow \theta - \eta \nabla_\theta \mathcal{L}_{total}(\theta)$
\EndFor
\State \textbf{return} Trained S$^2$Drug model
\end{algorithmic}
\end{algorithm*}

\end{document}